\documentclass[letterpaper, 10 pt, conference]{ieeeconf}

\IEEEoverridecommandlockouts

\usepackage{color}
\usepackage{xcolor}
\usepackage{epsfig}
\usepackage{graphicx}
\usepackage{algorithm,algorithmic}
\usepackage{censor}
\usepackage{dblfloatfix}

\usepackage{adjustbox}
\usepackage{array}
\usepackage{booktabs}
\usepackage{colortbl}
\usepackage{float,wrapfig}
\usepackage{framed}
\usepackage{hhline}
\usepackage{multirow}
\usepackage{makecell}
\usepackage[percent]{overpic}

\usepackage{amsmath,amsfonts,amssymb}
\usepackage{amsthm} 
\usepackage{bm}
\usepackage{nicefrac}
\usepackage[protrusion=true,expansion=true]{microtype}
\usepackage{contour}
\usepackage{courier}
\usepackage{siunitx}

\usepackage{changepage}
\usepackage{extramarks}
\usepackage{fancyhdr}
\usepackage{lastpage}
\usepackage{setspace}
\usepackage{soul}
\usepackage{xspace}
\usepackage{fancybox}
\usepackage{afterpage}
\usepackage{gensymb}

\usepackage{cite}
\usepackage{url}
\usepackage{hyperref}
\definecolor{MSBlue}{rgb}{0, 0.35, 0.49}
\hypersetup{colorlinks,linkcolor={black},citecolor={MSBlue},urlcolor={magenta}}
\usepackage{quoting}
\usepackage{epigraph}

\usepackage{enumerate}
\usepackage{paralist,tabularx}
\usepackage{comment}
\usepackage{pdfpages}
\usepackage{caption}  

\usepackage{pifont}

\usepackage{MnSymbol}

\usepackage{bbold}

\let\labelindent\relax
\usepackage{enumitem}
\usepackage[para]{footmisc}

\makeatletter
\DeclareRobustCommand\onedot{\futurelet\@let@token\@onedot}
\def\@onedot{\ifx\@let@token.\else.\null\fi\xspace}

\makeatother

\definecolor{MyDarkBlue}{rgb}{0,0.08,1}
\definecolor{MyDarkGreen}{rgb}{0.02,0.6,0.02}
\definecolor{MyDarkRed}{rgb}{0.8,0.02,0.02}
\definecolor{MyDarkOrange}{rgb}{0.40,0.2,0.02}
\definecolor{MyPurple}{RGB}{111,0,255}
\definecolor{MyRed}{rgb}{1.0,0.0,0.0}
\definecolor{MyGold}{rgb}{0.75,0.6,0.12}
\definecolor{MyDarkgray}{rgb}{0.66, 0.66, 0.66}
\definecolor{MyPink}{rgb}{1, 0.75, 0.79}
\definecolor{GreenStarColor}{rgb}{0.54, 0.84, 0.41}
\definecolor{MSBlue}{rgb}{0, 0.35, 0.49}
\definecolor{ditto}{RGB}{156,135,190}
\definecolor{nf}{RGB}{84, 156, 59}
\definecolor{ft}{RGB}{57, 118, 176}

\newcommand{\task}[1]{%
  \texttt{#1}%
}
\newcommand{\ditto}[1]{%
  \textcolor{ditto}{\texttt{#1}}
}
\newcommand{\nf}[1]{%
  \textcolor{nf}{\texttt{#1}}
}
\newcommand{\ft}[1]{%
  \textcolor{ft}{\texttt{#1}}
}

\def\OURS{DITTO\xspace}

\newcommand{\cmark}{\textcolor{green!60!black}{\ding{51}}}
\newcommand{\xmark}{\textcolor{red}{\ding{55}}}

\title{\LARGE \bf
\OURS: Dexterous Interface for Transparent TeleOperation}

\author{Joaquin Palacios$^{*,1}$,
Katelyn Lee$^{*,1}$,
Cheng Zhang$^{2}$,
Zhanpeng He$^{\dagger,3}$, and
Matei Ciocarlie$^{\dagger,1}$}

\begin{document}

\twocolumn[{%
\renewcommand\twocolumn[1][]{#1}%
\maketitle
\begin{center}
    \centering
    \vspace{-1em}
    \captionsetup{type=figure, font=small}
    \includegraphics[
        width=1.0\textwidth,
        trim=0cm 0cm 0cm 0cm,
        clip
    ]{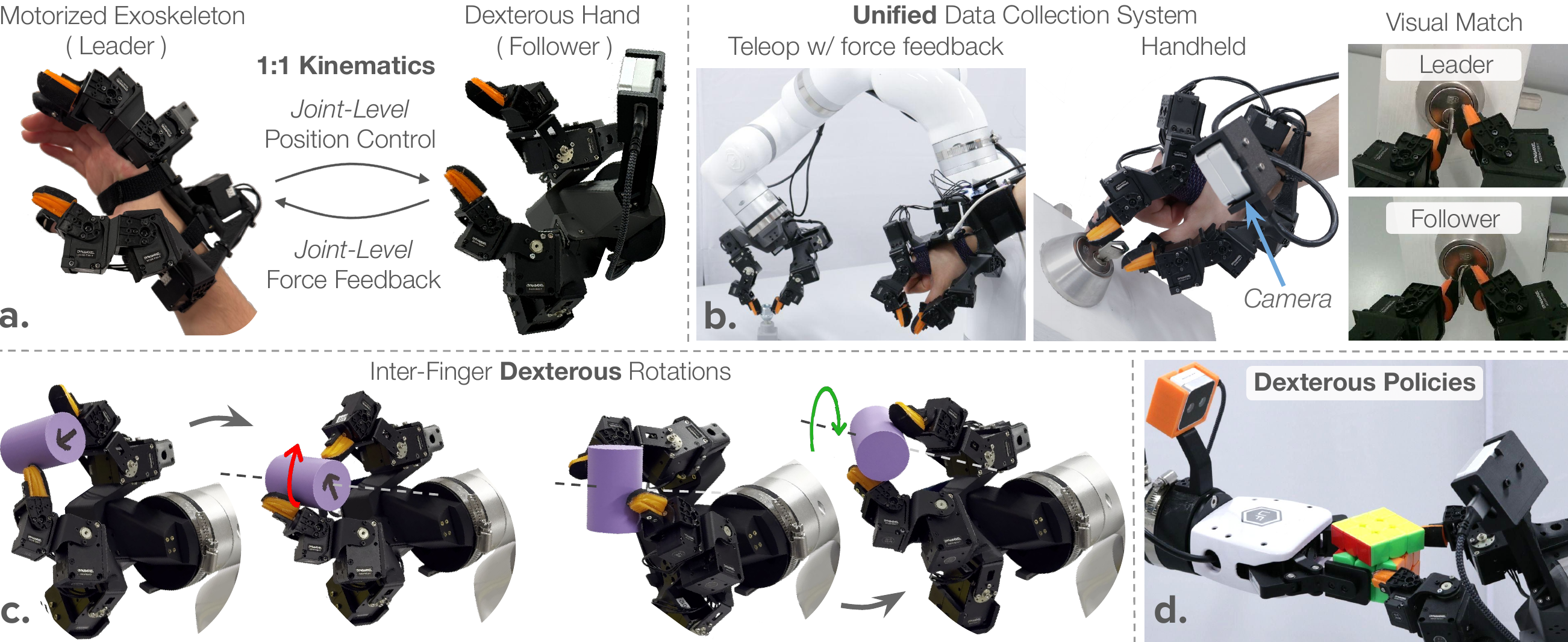}
    \captionof{figure}{\textbf{Overview of \OURS.}  (a) 1-to-1 leader-follower design enables \textit{joint-level} force feedback and allows for (b) a unified platform supporting both \textit{bilateral teleoperation} and \textit{handheld} (in-the-wild) data collection. DITTO's anatomy-based design is capable of (d) human-like dexterous primitives, such as multi-axes rotations. We demonstrate dexterity and data quality via (e) contact-rich manipulation policies. Videos at: https://ditto-robotics.github.io/. 
    }
    \label{fig:eyecandy}
\end{center}
}]

{
  \renewcommand{\thefootnote}%
    {\fnsymbol{footnote}}
  \footnotetext[1]{Equal contribution.} \footnotetext[2]{Equal advising.}
  \footnotetext[0]{$^{1}$Dept.\ of Mechanical Engineering, $^{2}$Dept.\ of Computer Science, Columbia University. $^{3}$Dept.\ of Computer Science, Stanford University.}
}

\thispagestyle{empty}
\pagestyle{empty}

\begin{abstract}
Collecting for manipulation with high-DOF hands is challenging, as interfaces must capture rich hand motion while rendering the contact interactions essential for precise manipulation. Existing data collection approaches face a trade-off: teleoperation ensures deployment consistency but lacks force feedback, while handheld (in-the-wild) systems provide natural force transparency but introduce a visual embodiment gap at deployment. We present DITTO, a Dexterous Interface for Transparent TeleOperation, which resolves this through the anatomically informed co-design of a dexterous 7-DOF robotic hand and a kinematically equivalent motorized exoskeleton. A 1-to-1 actuator mapping between the exoskeleton and robotic hand enables handheld (in-the-wild) data collection and bilateral teleoperation with \textit{joint-level} force feedback unified in a single platform. We demonstrate that the DITTO exoskeleton spans the operator's natural index-to-thumb workspace, and showcase DITTO's dexterous capabilities via learned policies on contact-rich tasks. Videos are available at: https://ditto-robotics.github.io/.

\end{abstract}


\section{Introduction}
\label{sec:intro}

Human demonstrations are a critical source of data for learning robot manipulation policies~\cite{intelligence2025pi, bjorck2025gr00t}. In particular, two dominant paradigms have emerged for collecting manipulation demonstrations: \textit{teleoperation} and \textit{handheld (in-the-wild)} collection. However, both paradigms remain challenging for highly articulated hands that use intrinsic DOFs for in-hand object reorientation and fine-grained regulation of contact forces between the hand, objects, and environment.

In \textit{teleoperation}, human movement is recorded and used to command a remote robot manipulator to perform a task for data collection. This ensures that, from the perspective of the robot, demonstration data matches autonomous deployment. However, common teleoperation interfaces such as vision-based tracking~\cite{qin2023anyteleop} and sensor gloves~\cite{dexterity_gen} can make fine-grained control unintuitive due to lack of force feedback and the kinematic gap between the operator's hand and the robot. Wearable motorized exoskeletons are an attractive alternative as they can provide force feedback at the fingertips~\cite{doglove, gex}, but they do not resolve this kinematic gap.

In contrast, in the \textit{handheld} or \textit{in-the-wild} data collection paradigm, the demonstrator collects manipulation data directly through an interface physically attached to their hand~\cite{dexumi, fang2025dexop, zhu2026dexexowearabilityfirstdexterousexoskeleton}. This mode is scalable and intuitive: the demonstrator receives direct force feedback from physical contact, and the interface itself constrains motion to the target hand's kinematics. However, it introduces a gap between data collection and deployment in both observation and action. For palm-mounted cameras, visual differences can be mitigated through in-painting ~\cite{dexumi} or hardware appearance matching ~\cite{fang2025dexop, zhu2026dexexowearabilityfirstdexterousexoskeleton}, but images from extrinsic cameras still include the operator, creating a visual mismatch. Moreover, because the robot's actuators do not execute the demonstration, the collected actions do not faithfully capture contact-rich interactions. Lastly, although mechanically constraining the user to the target hand's kinematics reduces the kinematic gap, if the robot's kinematics are not well aligned with the user's natural workspace, it can hinder their dexterity and natural motion.

These paradigms expose a trade-off: \textit{teleoperation} collects high-fidelity, robot-native data, but high-DOF control is difficult due to a lack of kinematic and force transparency. \textit{In-the-wild} interfaces provide intuitive collection with natural force feedback, but introduce mismatches between data collection and deployment in both observation and action.

We present \OURS, \textbf{D}exterous \textbf{I}nterface for \textbf{T}ransparent \textbf{T}ele\textbf{O}peration, which bridges this trade-off through the anatomically informed co-design of a dexterous 7-DOF 2-finger (thumb--index) \textbf{robot hand} and a matching \textbf{motorized and wearable exoskeleton} (Figure \ref{fig:eyecandy}). The key features of DITTO, and the main contributions of this paper, include:
\begin{itemize}[leftmargin=*]
    \item With only two fingers, DITTO uses a high-DOF kinematic structure to provide in-hand dexterity using intrinsic finger motion, without need for re-grasping or large wrist movement. In particular, our kinematic analysis on human manipulation data shows that DITTO's workspace spans the natural human index-to-thumb dexterous workspace.
    \item DITTO provides a 1:1 kinematic and actuator correspondence between the exoskeleton and its corresponding robot hand. This co-design enables \textbf{joint-level force feedback} for force-aware teleoperation even for dexterous primitives such as continuous rotation and in-hand object reorientation.
    \item DITTO provides a unified platform supporting both scalable handheld data collection \textbf{and} bilateral teleoperation with force feedback, enabling complementary data sources within a single system. Our user study (n=8) shows that force-aware teleoperation produces faster and better demonstrations (lower unwanted forces), and that established methods such as Diffusion Policies can readily learn from both types of DITTO data across contact-rich tasks that leverage intrinsic dexterity.
\end{itemize}

\section{Related Work}
\label{sec:related}

\textbf{Teleoperation for dexterous hands:} Controlling high-DOF hands for dexterous tasks requires 1) fine-grained control and 2) regulating contact forces during manipulation. The former is challenging due to the kinematic gap between the operator's and the robot's hand; the latter is challenging due the lack of force feedback, relying purely on vision. Commonly used motion tracking-based systems ~\cite{qin2023anyteleop} are lightweight and easy to use for simple grasping tasks, but are limited by these challenges in dexterous tasks.

Leader-follower designs like TILDE ~\cite{tilde} can close the kinematic gap directly through joint-level correspondence. However, TILDE is not wearable, making arm control during teleoperation difficult, and lacks force feedback. A complementary line of work narrows the kinematic gap in software rather than hardware, augmenting kinematic retargeting with reinforcement-learned skill priors~\cite{dexterity_gen}; this can improve teleoperation fidelity but requires a trained skill prior and still provides no force transparency.

Motorized hand exoskeletons aim to improve force regulation by rendering force feedback to the operator during teleoperation~\cite{doglove, gex, n2d}. However, in packaging actuators for force feedback, these devices forgo kinematic correspondence. This means they also rely on kinematic retargeting for motion. This limitation extends to force feedback, where devices either simplify feedback to a single-axis force per finger (i.e. DOGlove), or retarget forces in task space (i.e GEX~\cite{gex} and N2D~\cite{n2d}), both of which do not provide understanding of which joint is applying/experiencing forces.

\begin{table}[t]
\centering
\scriptsize
\setlength{\tabcolsep}{3.5pt}
\caption{Comparison with previous devices.}
\vspace{-4.5pt}
\label{tab:comparison}
\begin{tabular}{lccccc}
\multicolumn{1}{c}{Device}
&
\multicolumn{1}{c}{Tasks} &
\multicolumn{1}{c}{Teleoperation} &
\multicolumn{2}{c}{In-The-Wild} &
\multicolumn{1}{c}{Observation} \\
&
IHD &
Force Feedback &
Supported &
NIP &
Ex. Cam. \\
\midrule
DOGlove~\cite{doglove}
    & \xmark & Single-Axis & \xmark & N/A & \cmark \\
GEX~\cite{gex}
    & \xmark & Task-Space & \xmark & N/A & \cmark \\
DexUMI~\cite{dexumi}
    & \cmark & \xmark & \cmark & \xmark & \xmark \\
DEXOP~\cite{fang2025dexop}
    & \cmark & \xmark & \cmark & \cmark & \xmark \\
DEX-Mouse~\cite{dexmouse}
    & \xmark & Single-Axis & \cmark & \cmark & \xmark \\
\textbf{DITTO (ours)}
    & \cmark & \textbf{Joint-Level} & \cmark & \cmark & \cmark \\
\bottomrule
\end{tabular}

\vspace{4pt}
\parbox{\columnwidth}{
\scriptsize
IHD = in-hand dexterity, or ability to reorient objects with intrinsic DOFs; NIP = no in-painting needed; Ex. Cam.\ = extrinsic camera without human operators in the frame.
}
\vspace{-3em}
\end{table}

\textbf{Handheld (In-the-wild) data collection:} Handheld devices mitigate the issues of teleoperation by having users perform the task via a wearable data collection interface, directly feeling interaction forces during manipulation. DexUMI~\cite{dexumi} is a wearable exoskeleton that constrains the user's hand to the target kinematics, but requires in-painting to reduce the visual mismatch compared to the target hand. DEXOP~\cite{fang2025dexop} and DexEXO~\cite{zhu2026dexexowearabilityfirstdexterousexoskeleton} instead use linkages connecting the user's fingertips to the target hand, positioning the palm camera to capture the actual robot and reduce the visual mismatch (in the palm camera), eliminating the need for in-painting.

However, embodiment gaps remain in two forms: 1) external camera views still include the operator in frame, creating a large visual mismatch (which does not exist during teleoperation), and 2) the user performs the actions and applies forces, so demonstrations do not capture the robot's actuator dynamics. DEX-Mouse ~\cite{dexmouse} is a hybrid of teleoperation and handheld collection system where the user wears both the target hand and a motorized wearable, controlling the hand's wrist pose with their own arm and teleoperating the robot's finger motion to fully capture the hand's actuator dynamics. However, this mode still does not support extrinsic camera views without visual mismatch, and is limited in dexterity due to one DOF and actuator per finger.

From a kinematic standpoint, this family of devices constrain the user's hand movements to the target hand's kinematics, which can constrain the human hand's natural workspace. DEXOP ~\cite{fang2025dexop} mitigates the workspace constraint by co-designing the wearable interface and robotic hand, but does not provide quantitative kinematic analysis of how well the workspace aligns with human hand.

Table~\ref{tab:comparison} places DITTO in the context of existing devices. DITTO co-designs a wearable, \textit{motorized} exoskeleton with a matching 7-DOF thumb--index follower hand. The resulting 1:1 actuator mapping enables joint-level force feedback during bilateral teleoperation, supports handheld in-the-wild collection without in-painting, and, in teleop mode, allows extrinsic cameras without the operator in frame, while retaining in-hand dexterity via intrinsic finger motion.

\section{DITTO Hardware Design}
\label{sec:method}

\begin{figure}[t]
    \centering
    \includegraphics[width=\columnwidth]{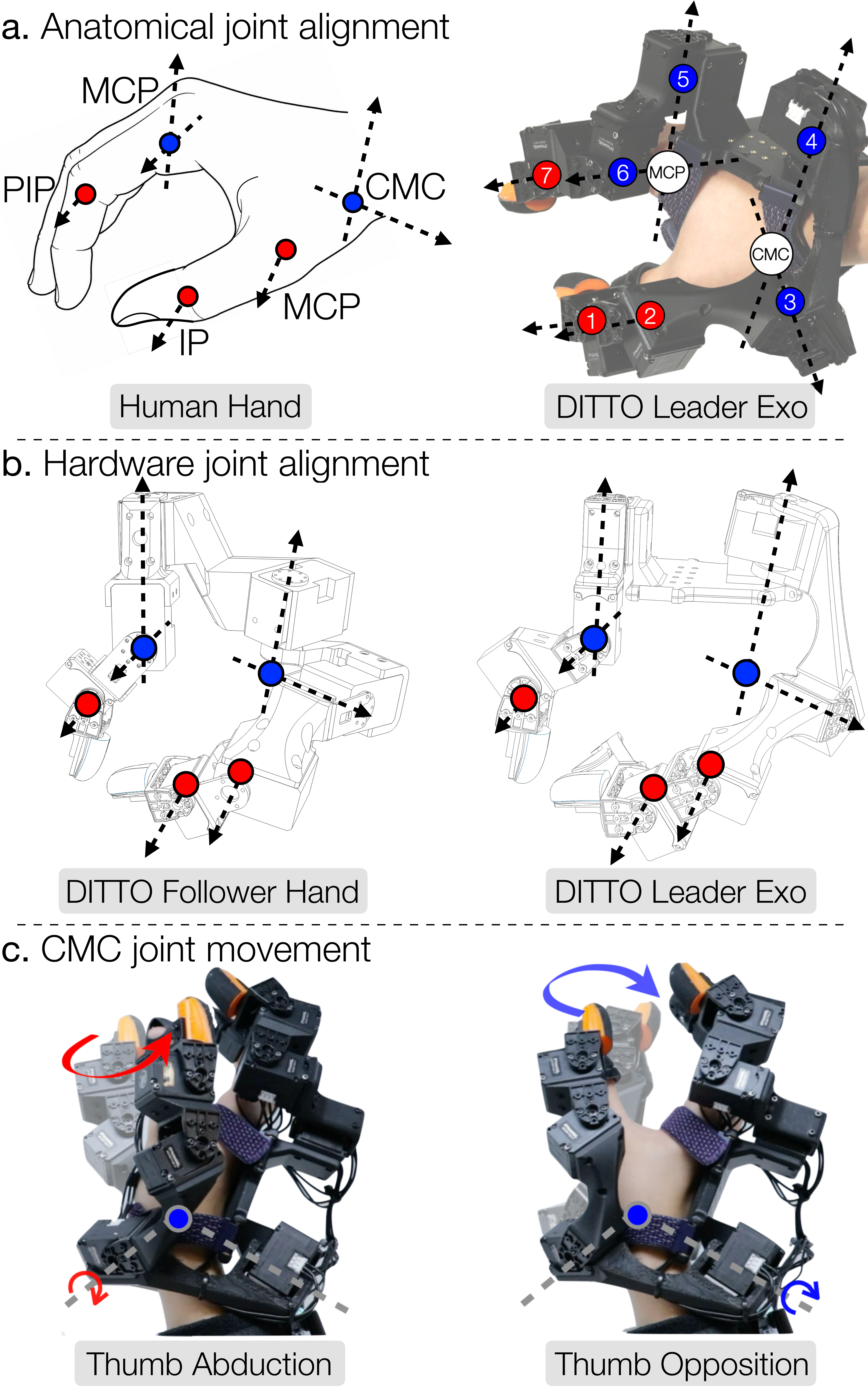}
    \caption{\label{fig:leader_follower_kinematics} (a) \OURS aligns each of its 7 motor axes with a finger joint axis. In blue, the 2-DoF CMC and MCP joints have offset, decoupled motors so the motor axes intersect at the anatomical joint center. (b) \OURS has 1:1 kinematic equivalence between the leader exoskeleton and the follower hand. (c) The offset, decoupled motors for the CMC enable dexterous thumb motions like abduction and opposition.}
\end{figure}

DITTO aims to bridge the fundamental trade-off between kinematic and force transparency in teleoperation by co-designing a dexterous 7-DOF hand and a kinematically equivalent motorized wearable exoskeleton. This leader-follower system has a 1:1 actuator mapping, where every motor on the follower hand corresponds to a motor on the leader exoskeleton. This concept has been demonstrated for arm teleoperation \cite{gello, aloha} but not for high-DOF hands in a wearable form factor. The primary difficulty lies in the non-obtrusive, wearable packaging of enough actuators to achieve complex tip-to-tip movement without restricting the user's natural workspace. In this section, we describe how we achieve this via careful, anatomically-inspired kinematic co-design and how the 1:1 actuator mapping enables joint-level force rendering and transparency during teleoperation.

\subsection{DITTO Kinematics}
\label{subsec:kinematics}
\paragraph{Leader Kinematics}
We designed DITTO around the human hand, using anatomical finger kinematics as the basis for both the wearable exoskeleton (leader) and the robotic hand (follower). For the exoskeleton, we prioritize wearability and anatomical joint alignment, ensuring the exoskeleton axes coincide with the biological joint axes of the index finger and thumb. The DITTO index finger has three DOF: one aligning with the PIP (proximal interphalangeal) joint, and two aligning with the MCP (metacarpophalangeal) joint. The DITTO thumb has four DOF: two aligning with the IP (interphalangeal) and MCP flexion joints, and two aligning with the CMC (carpometacarpal) joint. The CMC joint is a complex biological joint whose motion is commonly approximated as a two DOF universal joint with perpendicular axes, and we adopt this simplification in our design \cite{cmc_kinematics}.

To accommodate the servo motors within the form factor of a wearable hand exoskeleton, we position them along the plane formed by the tip-to-tip pinch grasp between the index finger and thumb (Fig.~\ref{fig:leader_follower_kinematics}a.). For the index finger PIP joint and the thumb flexion degrees of freedom, this placement is straightforward, as the joint axes align naturally with motor placement in this plane. For the index MCP and thumb CMC joints—each of which we model as a two DOF universal joint—direct motor placement is more constrained. We address this with a non-anthropomorphic design that decouples each universal joint into two offset joints whose axes intersect at the anatomical center of rotation of the human joint (Fig.~\ref{fig:leader_follower_kinematics}c.). This decoupled arrangement allows each motor to align with one rotational axis of the universal joint while remaining positioned around the hand.

The exoskeleton attaches to the operator's hand at three points of contact: a rigid handpiece on the dorsal side of the hand and two finger caps on the distal links of the index finger and thumb. Users don the exoskeleton by aligning with their index MCP joint and securing the base to the dorsal side of their hand. Each fingercap consists of a silicone cap and a rigid piece that connects the user's fingertips: one attaches at the PIP joint on the index finger frame, and the other at the IP joint on the thumb frame. To accommodate a range of hand sizes, these finger interface pieces can be customized to different finger lengths.

\paragraph{Follower Kinematics}
We derive the robot follower hand kinematics directly from the exoskeleton leader to keep the same joint positions and axes for 1:1 kinematic transparency. Without the space constraint of the operator's hand on the exoskeleton leader, we can use larger servo motors with greater output torque for the robot follower hand. The larger motors impose space constraints and require two joints to be translated along the joint axis: the index MCP abduction joint by 12 mm and the first thumb CMC joint by 33.1 mm. All remaining links are kinematically identical to the leader, preserving joint-level transparency between the operator and the follower hand.

\begin{figure*}[t]
    \centering
    \includegraphics[width=0.9\textwidth]{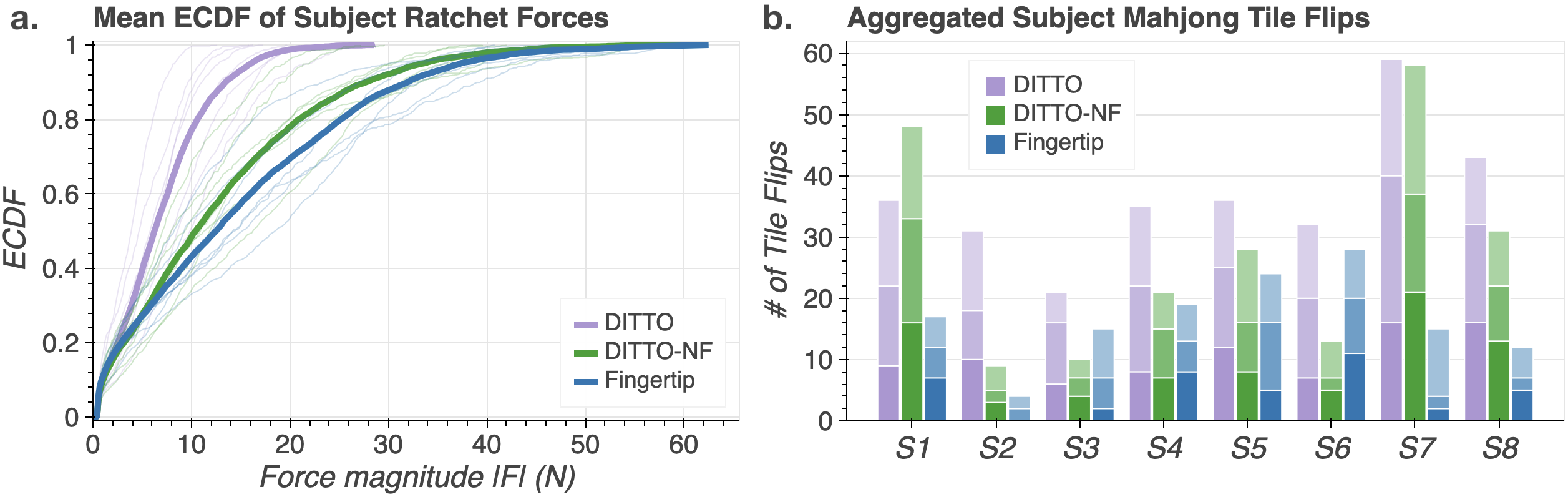}
    \caption{(a) The three bolded lines in the ECDF plot denote the gross average over all $n=8$ subjects per teleop condition, and the faint lines denote the individual subject curves. (b) The stacked bar plot tallies the total number of mahjong tile flips over the three trials, in order, with teleop condition labeled by color and trial number denoted by opacity.}
    \label{fig:user_study_figs}
\end{figure*}

\subsection{Hardware characteristics}
\label{subsec:hardware_characteristics}

The leader exoskeleton is low-cost at $\approx$\$900 and weighs 430\,g; donning and doffing take about 20\,s and 10\,s, respectively. The follower hand weighs 935\,g and costs $\approx$\$1500, a substantially lower price than typical commercial dexterous hands. We selected Dynamixel XC330-T181 motors for the leader exoskeleton for their small size and because their 181:1 gear ratio was small enough to allow users to comfortably backdrive them during motion capture. The follower hand uses Dynamixel XM430 motors for all joints except the most distal index PIP and thumb IP joints, which use smaller Dynamixel XC330-T288 motors.

\subsection{Teleoperation with Force Feedback}
\label{subsec:force_rendering}

DITTO's leader/follower 1:1 actuator mapping enables joint-level force feedback with direct per-joint correspondence. This has the added benefit of reducing bilateral teleoperation to simple per-joint loops:

\begin{itemize}[leftmargin=*]
    \item \textbf{Leader to follower}: Joint-level position tracking via Dynamixel position control.
    \item \textbf{Follower to leader}: Force feedback on the leader is computed independently for each joint using the corresponding follower motor current. A deadband is applied to suppress feedback during free-space motion:
\begin{equation}
\tau_l =
\begin{cases}
- k_f \left( \tau_f - \tau_{\text{th}} \right), & \text{if } |\tau_f| > \tau_{\text{th}} \\
0, & \text{otherwise}
\end{cases}
\end{equation}
where $\tau_f$ denotes the estimated torque from motor current of the follower actuator, $k_f$ is a feedback gain, and $\tau_{\text{th}}$ defines a deadband threshold that filters out low-magnitude currents associated with general motion, while capturing high-magnitude currents from contacts and applied forces.
\end{itemize}

This structure eliminates the need for task-space force mapping and reduces computational overhead. Our system supports bilateral teleoperation at 200 Hz using off-the-shelf U2D2's for communication with Dynamixel motors. User safety is enforced via conservative joint torque limits ($200\,\mathrm{mA} \approx 0.08\,\mathrm{Nm}$), position limits, and an emergency stop.

\subsection{Handheld Data Collection}
\label{subsec:handheld_hardware}

While \OURS can be used for leader-follower teleoperation, the \OURS exoskeleton by itself can also be used directly for handheld data collection.
To this end, the \OURS leader is equipped with a camera and a pose tracking sensor. The camera positioning matches that of camera on the follower hand and is carefully placed to ensure only the exoskeleton is visible and not the user's hand, ensuring minimal visual gap  without requiring post-processing or in-painting (Fig. \ref{fig:eyecandy}.b).

\section{User Study: Dexterous Data Collection}

When used in leader-follower teleoperation mode, \OURS aims to provide active force feedback while still allowing rich tip-to-tip movement for dexterous tasks. To assess the importance of these features, we conducted a user study ($n=8$) comparing DITTO teleoperation against two ablations and across two tasks that require coordinated fingertip dexterity and force regulation. We compare three \textbf{teleoperation conditions}:
\begin{itemize}[leftmargin=*]
    \item \ditto{DITTO} exoskeleton: leader-follower teleoperation using \OURS exoskeleton with force feedback enabled.
    \item \nf{DITTO-NF} (No Force): an ablation of DITTO's bilateral force rendering. The demonstrator uses the DITTO leader exoskeleton joint position commands are related to the follower hand, but force feedback on the leader is disabled.
    \item \ft{Fingertip}: an ablation of DITTO's joint-level kinematic correspondence. The leader exoskeleton is not used at all; instead, we attach 6-DOF pose trackers to the operator's fingertips and retarget fingertip motion to the DITTO follower hand via inverse kinematics.
\end{itemize}

For each teleoperation condition, subjects completed two \textbf{tasks}: \task{mahjong tile} and \task{ratchet}. In the \task{mahjong tile} task, subjects flip a mahjong tile as many times as possible within a 60-second interval. The tile rests on a platform mounted to a 6-axis force-torque sensor, and subjects teleoperate a 7-DOF robot arm while using the DITTO follower hand to manipulate the tile. A successful flip is defined as a clockwise, 90 degree rotation about the tile's long axis. This task evaluates speed in fingertip dexterity, as successful manipulation requires coordinated index-thumb motion to quickly grasp and rotate the tile. In the \task{ratchet} task, subjects rotate a ratchet for three full revolutions as fast as possible. The 7-DOF robot arm is locked into a fixed pose relative to the ratchet to isolate interaction forces to the DITTO follower hand to evaluate force regulation and whether bilateral force rendering reduces excessive spurious forces during manipulation.

\textbf{Protocol} All $n=8$ subjects were recruited via our IRB-approved protocol and had no prior experience with hand teleoperation. To mitigate order effects across teleoperation conditions and minimize exoskeleton donning and doffing time, we used a hierarchical randomization procedure via coin flip. For each participant, we first randomized whether trials began with \ft{Fingertip} tracking or a DITTO-based condition. The order of the two DITTO-based conditions, \nf{DITTO-NF} and \ditto{DITTO}, was then randomized independently. The order of tasks (\task{mahjong tile},\ \task{BOA}) was kept constant to ensure that any task learning effects were consistent across participants and conditions.

\begin{table}[t]
    \vspace{-1pt}
    \centering
    \caption{\label{tab:user_study_summary} User study metrics (avg \& std for 8 users).}
    \setlength{\tabcolsep}{3pt}
    \begin{tabular}{@{}r|c|c|c@{}}
    \task{task} \hspace{1.4em} metric & \ditto{DITTO} & \nf{DITTO-NF} & \ft{Fingertip} \\[1pt]
    \hline
    \task{mahjong tile} & & & \\[1pt]
    \hspace{1.5em}flips (\#)
        & $\mathbf{36.6 \pm 11.0}$ & $27.2 \pm 18.0$ & $16.8 \pm 7.3$ \\[2pt]
    \task{ratchet\hspace{3.1em}} & & & \\[1pt]
    \hspace{1.5em}time (s)
        & $24.1 \pm 6.6$ & $\mathbf{22.4 \pm 9.1}$ & $34.2 \pm 8.1$ \\
    \hspace{1.5em}force (N)
        & $\mathbf{6.9 \pm 2.0}$ & $13.0 \pm 3.0$ & $14.8 \pm 2.5$ \\
    \hspace{2.5em}\% $\leq$ 14.8N
        & $\mathbf{92.8 \pm 8.7}$ & $64.8 \pm 11.5$ & $56.7 \pm 10.7$ \\
    \hline
    \end{tabular}
    \vspace{-15pt}
\end{table}

\textbf{Metrics}\quad We evaluated performance using four metrics \ref{tab:user_study_summary}. The total flips (\#) is the throughput of the number of successful flips completed within the 60-second trial on the \task{mahjong tile} task, aggregated over three trials per condition, capturing the manipulation efficiency of each teleoperation condition. For the \task{ratchet} task, we measure task completion time, defined as the total time required to complete three full revolutions. We also measure the interaction force magnitude with a 6-axis f/t sensor mounted beneath the ratchet; lower values indicate finer force control.

To characterize the force distribution during active manipulation in the \task{ratchet} condition, we aggregated all trials per condition and computed the empirical distribution function (ECDF), $F(x) = \frac{\#\text{ of points } \leq x}{n}$, where $x$ is the force magnitude, and $n$ is the total number of samples in the condition. We can define a reference threshold at $14.8$ N, the average force value from the \ft{Fingertip} condition [Table~\ref{tab:user_study_summary}], and calculate the ECDF value for each condition, denoted as \% $\leq$ 14.8N.

\textbf{Statistical significance}\quad One-sided permutation tests found statistically significant differences for \ditto{DITTO} vs. \ft{Fingertip} across all metrics (all $p \leq$ 0.023) and for \ditto{DITTO} vs. \nf{DITTO-NF} for the two \task{ratchet} force metrics ($p =$ 0.012). \task{ratchet} times for \nf{DITTO-NF} are slightly faster than \ditto{DITTO}, but elevated \task{ratchet} forces suggest that users, without force feedback, speed up task completion with more aggressive behavior.

\textbf{\task{Mahjong Tile} results}\quad Aggregating across all three trials, every subject completed the highest number of successful flips with the DITTO exoskeleton, and all but S1 achieved their highest total with force rendering enabled [Fig.~\ref{fig:user_study_figs}a.]. On average, subjects completed $36.6 \pm 11.0$ total flips using \ditto{DITTO}, compared to $27.28 \pm 18.0$ with \nf{DITTO-NF} and more than $2 \times$ the $16.8 \pm 7.3$ total flips under \ft{Fingertip} \ref{tab:user_study_summary}. The gap between \ft{Fingertip} and the two exoskeleton conditions indicates that DITTO's kinematic structure supports more effective coordinated fingertip manipulation than fingertip retargeting alone.

We defined two failure cases for the \task{mahjong tile} task: the tile getting flicked off of the platform and overloading of the force-torque sensor due to excessive contact forces. These failures occurred in $6/8$ subjects, and only occurred under the \ft{Fingertip} and \nf{DITTO-NF} conditions. For S2, all \ft{Fingertip} trials and 2 of 3 \nf{DITTO-NF} trials failed due to sensor overloading (force values of $\geq 100$~N). These failure cases suggest that force feedback also helps subjects regulate object interaction forces during manipulation, preventing excessive forces that could damage the manipulated object or exceed sensor limits.

\textbf{\task{Ratchet} results}\quad For the \task{ratchet} task, \ditto{DITTO} yielded the lowest average interaction forces for every subject at $6.9 \pm 2.0$~N, compared to $13.0 \pm 3.0$~N under \nf{DITTO-NF} and $14.8 \pm 2.5$~N under \ft{Fingertip} [Table~\ref{tab:user_study_summary}]. We filter out force magnitudes below 0.45 N (3$\times$ the 0.15N resolution of the force-torque sensor) to exclude idle periods between active manipulation, which can bias the average force downward and obscure differences in force regulation during the task.

The ECDF plots show that \ditto{DITTO} samples are concentrated at lower forces compared to the other teleoperation conditions, with the \ditto{DITTO} lines having a steeper and shorter curve than \nf{DITTO-NF} and \ft{Fingertip} lines [Fig~\ref{fig:user_study_figs}]. Notably, Table~\ref{tab:user_study_summary} shows that for all subjects, \ditto{DITTO} has the largest percentage of force values below the threshold, indicating that force feedback reduces the prevalence of high-force interactions.

While \nf{DITTO-NF} achieved the fastest average completion times, these trials also exhibited higher interaction forces, suggesting that without force feedback, operators applied more aggressive forces to complete the task quickly. This indicates that the higher forces under \nf{DITTO-NF} and \ft{Fingertip} were \textit{not} essential for efficient performance, but rather a consequence of \textit{reduced force awareness} during teleoperation.

\section{Kinematic Workspace Analysis}

Matching human index-to-thumb dexterity is a key design goal for \OURS providing two important benefits. First, from the perspective of the \OURS leader (a wearable exoskeleton), a human-like workspace removes constraints on the teleoperator's hand to allow natural and intuitive movement without pressure points. Second, from a follower perspective, human-like tip-to-tip movement implies high dexterity, since the human hand can perform complex manipulation with just index-to-thumb grasps exclusively.

In order to assess the \OURS tip-to-tip workspace, and to compare it against the human hand, we performed a kinematic analysis against data collected from seven subjects of varying hand sizes. We use this analysis to quantitatively compare the \OURS leader exoskeleton both against the human thumb-to-index workspace and against two other high-DOF wearable exoskeletons: DOGlove~\cite{doglove} and DexUMI-Xhand exoskeleton~\cite{dexumi}.

\begin{table*}[b]
    \centering

    \resizebox{\textwidth}{!}{
    \begin{tabular}{lccccccccccccccc}
    \toprule
    & \multicolumn{3}{c}{\textbf{Ratchet}} & \multicolumn{3}{c}{\textbf{Cube}} & \multicolumn{3}{c}{\textbf{Cap}} & \multicolumn{3}{c}{\textbf{USB}} & \multicolumn{3}{c}{\textbf{Mahjong}} \\
    \cmidrule(lr){2-4}\cmidrule(lr){5-7}\cmidrule(lr){8-10}\cmidrule(lr){11-13}\cmidrule(lr){14-16}
    \textbf{System} & \textbf{Succ.} & \textbf{X-Err} & \textbf{Pos.} & \textbf{Succ.} & \textbf{X-Err} & \textbf{Pos.} & \textbf{Succ.} & \textbf{X-Err} & \textbf{Pos.} & \textbf{Succ.} & \textbf{X-Err} & \textbf{Pos.} & \textbf{Succ.} & \textbf{X-Err} & \textbf{Pos.} \\
    \midrule
    DITTO & \textbf{1.0$\pm$0.0} & \textbf{4.2$\pm$1.6} & \textbf{1.0$\pm$0.0} & \textbf{1.0$\pm$0.1} & \textbf{4.0$\pm$2.2} & \textbf{1.0$\pm$0.0} & \textbf{1.0$\pm$0.0} & \textbf{3.0$\pm$0.9} & \textbf{1.0$\pm$0.0} & \textbf{0.8$\pm$0.2} & \textbf{1.7$\pm$0.7} & 0.9$\pm$0.1 & \textbf{1.0$\pm$0.0} & \textbf{3.4$\pm$1.1} & \textbf{1.0$\pm$0.0} \\
    DexUMI & 0.1$\pm$0.2 & 32.1$\pm$8.1 & \textbf{1.0$\pm$0.0} & 0.3$\pm$0.3 & 28.2$\pm$9.3 & \textbf{1.0$\pm$0.0} & 0.6$\pm$0.4 & 17.8$\pm$7.9 & \textbf{1.0$\pm$0.0} & 0.1$\pm$0.1 & 31.2$\pm$10.7 & 0.4$\pm$0.1 & 0.2$\pm$0.3 & 30.6$\pm$12.5 & 0.8$\pm$0.1 \\
    Doglove & 0.5$\pm$0.2 & 12.1$\pm$5.4 & \textbf{1.0$\pm$0.0} & 0.5$\pm$0.3 & 13.5$\pm$7.5 & \textbf{1.0$\pm$0.0} & 0.3$\pm$0.3 & 18.4$\pm$7.3 & \textbf{1.0$\pm$0.0} & 0.1$\pm$0.2 & 21.2$\pm$6.1 & \textbf{1.0$\pm$0.0} & 0.2$\pm$0.3 & 18.6$\pm$6.0 & \textbf{1.0$\pm$0.0} \\
    \bottomrule
    \end{tabular}
    }
    \caption{Kinematic workspace coverage across five manipulation tasks. \textbf{Succ.} denotes the fraction of sampled poses satisfying the 1 mm position tolerance and orientation check. The orientation check requires the fingertip x-axis to lie within 20° of the target axis and remain in the correct yz- hemisphere. \textbf{X-Err} is the mean x-axis angular error, and \textbf{Pos.} is the position-only success rate. Values are mean, $\pm$std over all trials and participants ($n=7$). \textbf{Cube} averages four rotation directions; \textbf{Cap} averages cap and uncap trials.\label{tab:workspace_accuracy}}
\end{table*}

\textbf{Data collection:} We attach two 6-DOF motion capture sensors (trakSTAR, Ascension) to the index and thumb fingertips and record subjects performing several tasks that require coordinated index-thumb manipulation, capturing their relative range of motion during each task. Specifically, subjects completed eight tasks across three trials each: mahjong tile rotation, USB flash drive extrusion, ratchet rotation, test tube capping and uncapping, and clockwise/counterclockwise rotation of the top and right faces of a Rubik's cube. Because each subject's kinematic data defines the target poses, the reachable workspace depends directly on hand size. Across our seven subjects, index finger length had a range of 27 mm, palm width 31 mm, and thumb length 19 mm.

\begin{figure}
    \centering
    \includegraphics[
        width=1.0\linewidth,
        trim=0cm 0cm 0cm 0cm,
        clip
    ]{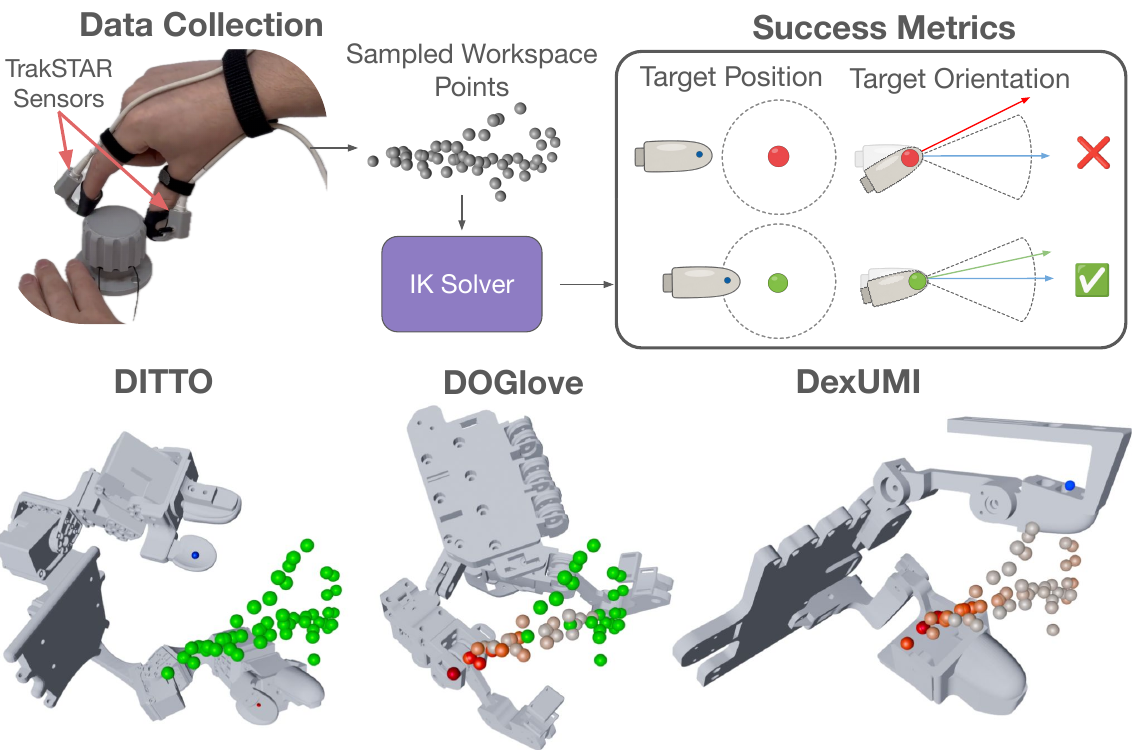}
    \caption{Top: Kinematic analysis data collection and data processing workflow. Bottom: Example workspace results visualized, targets colored by IK solution quality, from green (within tolerance) to red (error exceeds threshold).}
    \label{fig:ka_diagram}
\end{figure}

\textbf{Kinematic analysis:} After collecting the human kinematic data, we downsample the workspace using SE(3) Farthest Point Sampling (FPS) to get a subset of poses with near-uniform coverage of the original human workspace. Then, to assess relative pose of the index finger with respect to the thumb, we calculate the relative transform from the thumb tip to the index tip [Fig. \ref{fig:ka_diagram}].

For a given exoskeleton design, we are interested if the robot kinematics can achieve the same tip-to-tip relative poses as observed in the human data set. Thus, we take the kinematic chain of the exoskeleton, set the thumb tip as a fixed base, then compute inverse kinematics using mink IK~\cite{mink} for the index tip to check if a specific relative pose is feasible within a 1 mm position tolerance and 20$^{\circ}$ angular tolerance of the fingertip midline axis (defined as the x-axis). We then aggregate these results over the totality of the human tip-to-tip workspace and assess a robot's kinematic capability based on what percentage of human tip-to-tip poses it is capable of reproducing with low position and angular error.

\textbf{Results} are shown in Table~\ref{tab:workspace_accuracy} for \OURS as well as two other current teleoperation exoskeleton designs. They shows that \OURS reaches nearly the entire human thumb-to-index workspace, with a mean success rate at or near 1.0 and a mean x-axis orientation error of only 1.7$^\circ$ to 4.2$^\circ$. Within the same 1 mm position and 20$^\circ$ orientation bounds, DexUMI and DOGlove reach at most 0.6 and 0.5, respectively, with orientation errors 3$\times$ to 8$\times$ larger. All three systems achieve high position-only success rates, indicating that the success gap arises almost entirely from orientation: the anatomically aligned joint axes of the \OURS leader allow it to match both the position and orientation of natural fingertip poses. We believe this plays an important role in the fact that \OURS enables fast and high-quality data collection even for tasks that leverage intrinsic DOFs of the hand.

\begin{figure*}[t]
    \centering
    \includegraphics[width=0.85\textwidth]{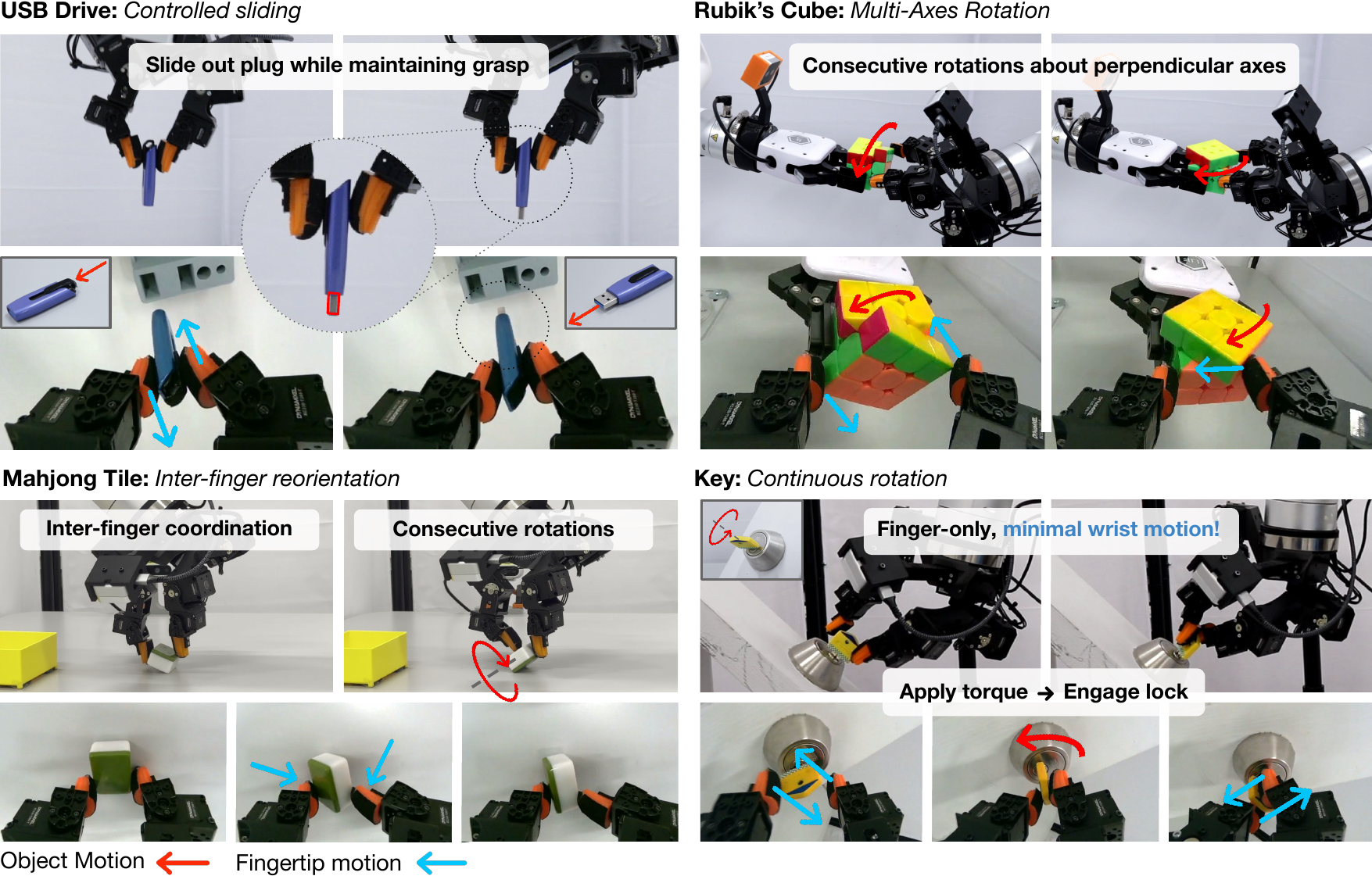}
    \caption{\label{fig:tasks_learning} Learned policies for dexterous and contact-rich manipulation tasks. For unimanual tasks, we include: \textbf{USB Drive Insertion}, which requires picking up and extending a retractable USB flash drive and then inserting it into a slot, \textbf{Mahjong Tile Flipping}, where the robot reorients a tile through two sequential flips followed by placement into a target box, and  \textbf{Key Rotation} to engage a lock. For the bimanual task, we include \textbf{Rubik's Cube Rotation}, which requires picking up and then rotating two faces of a Rubik's Cube using thumb-index finger manipulation. Videos at: https://ditto-robotics.github.io/}
\end{figure*}

\section{Policy Learning Results}
Finally, we evaluate weather DITTO data-collection enables autonomous policies for tasks requiring use of the intrinsic hand DOFs, as well as regulation of contact forces with the environment. We test these abilities for both data collection modes offered by DITTO, teleoperation and handheld, as well as combined datasets from both paradigms. For the autonomous policies, we use standard Diffusion Policies with a Transformer Architecture \cite{chi2025diffusion} using the relative pose representation from \cite{chi2024universal}.

\subsection{Policies Trained with Teleoperation Datasets}
\begin{table}[t]
    \centering
    \small
    \setlength{\tabcolsep}{4pt}
    \renewcommand{\arraystretch}{1.01}
    \begin{tabularx}{\columnwidth}{X c}
    \hline
    \multicolumn{2}{c}{\textbf{Unimanual Tasks}} \\
    \hline
    \textbf{Stage} & \textbf{Success} \\
    \hline
    \textbf{Key Rotation (60 Demos, 180 Epochs)} & \\
    Rotate key until lock engages & 20/20 \\
    \hline
    \textbf{Mahjong Flipping (56 Demos, 250 Epochs)} & \\
    Rotate twice & 20/20 \\
    Pick and place & 20/20 \\
    \hline
    \textbf{USB Insertion (55 Demos, 300 Epochs)} & \\
    Pick up USB & 20/20 \\
    Extend USB & 19/20 \\
    Insert into slot & 14/20 \\
    \hline
    \multicolumn{2}{c}{\textbf{Bimanual Tasks}} \\
    \hline
    \textbf{Stage} & \textbf{Success} \\
    \hline
    \textbf{Rubik's Cube (59 Demos, 300 Epochs)} & \\
    Rotate right face & 18/20 \\
    Rotate top face & 17/20 \\
    \hline
    \end{tabularx}
    \caption{\label{tab:teleop_tasks_success} Task success rate for policies trained on teleoperation data collected using DITTO.}
\end{table}

We evaluate manipulation performance across unimanual and bimanual dexterous tasks [Fig.~\ref{fig:tasks_learning}]. Our results in Table \ref{tab:teleop_tasks_success} show that policies trained on DITTO-collected data achieve high success rates across all tasks, showcasing the effectiveness of our data collection platform. Notably, the USB insertion task requires both in-hand manipulation and extrinsic contact reasoning, demonstrating that our device can capture the nuanced finger coordination needed for such complex interactions. On bimanual tasks, our policies also achieve high success rates, highlighting DITTO's ability to support coordinated two-hand manipulation and collect high-quality demonstrations beyond the unimanual setting.

\subsection{Policies Trained with Handheld and Combined Datasets}
To validate our handheld data collection capability, we evaluate and compare policies trained on \textit{handheld} data, \textit{teleoperated} data, and various mixtures of the two. We focus on a unimanual task requiring high finger-to-finger coordination: rotating a test tube cap to loosen it, then grasping and removing it without dropping it.
\begin{figure}[t]
    \centering
    \includegraphics[width=0.9\columnwidth]{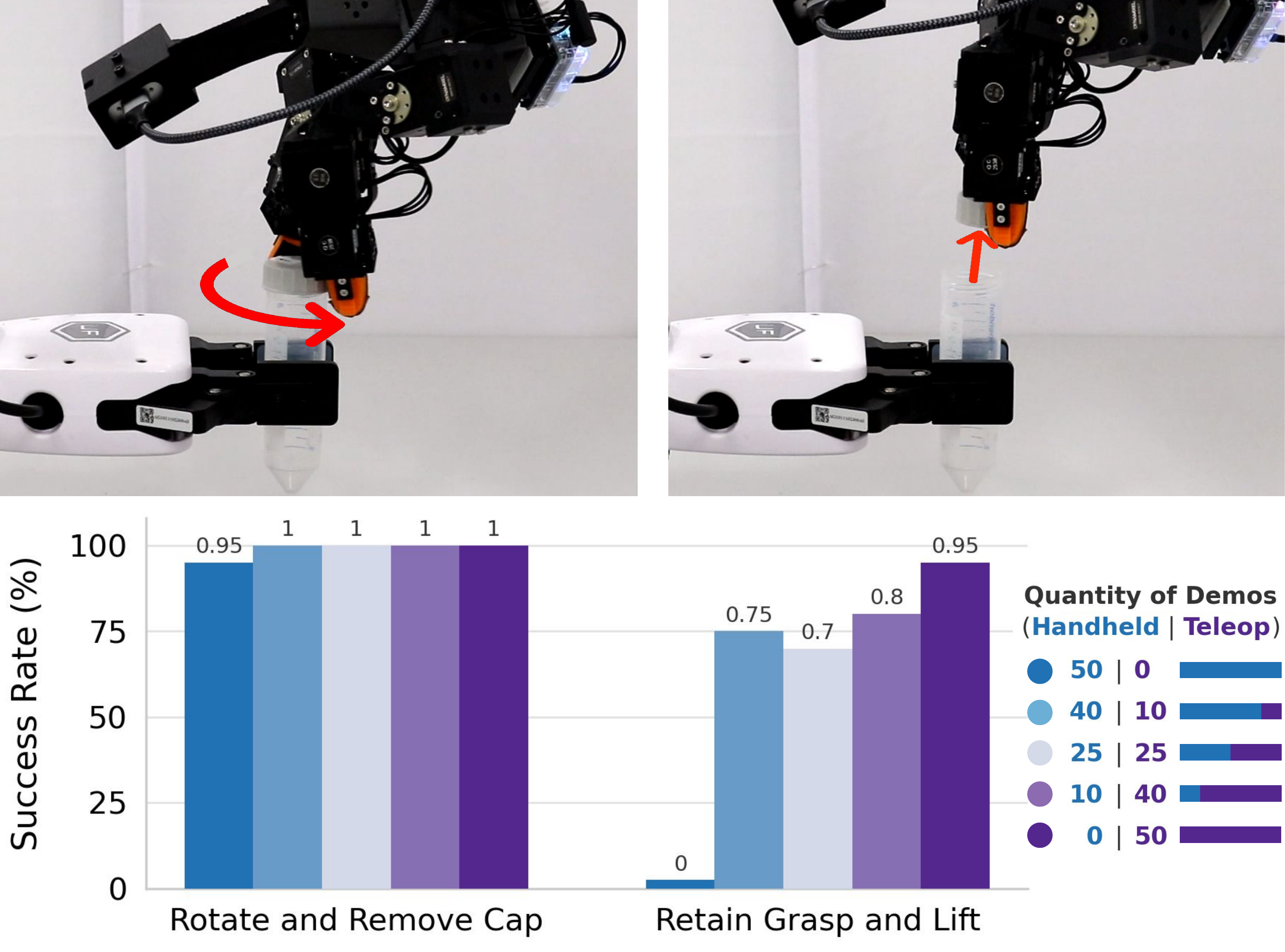}
    \caption{\label{fig:itw_teleop_uncap}Test tube uncapping policy performance for data mixing combinations of \textit{handheld} and \textit{teleoperation} demonstrations. All policies trained with 50 demos for 250 epochs.}

\vspace{-14pt}
\end{figure}

Our results show that for the rotate-and-remove subtask, the policy trained on \textit{handheld} data alone performs similarly to those mixed with \textit{teleoperation} data (small 5$\%$ margin of difference), showing that \textit{handheld} data captures the dexterous finger coordination needed for continuously rotating a cap. For the retain-grasp-and-lift subtask, however, policies trained on \textit{handheld} data alone fail entirely.  The primary failure mode is a poor grasp before lifting, which leads to dropping the object. Augmenting with a small amount of teleoperation data (10 episodes) raises success to 70\%, consistent with prior work claiming handheld data alone is insufficient and that some \textit{teleoperated} demonstrations are needed~\cite{fang2025dexop}. Training on teleoperation data alone further improves success by 20 percentage points. We hypothesize that the \textit{handheld}-only failure reflects an action mismatch: applying forces and movement is performed by the user rather than the robot's actuators, whereas teleoperated demonstrations record robot-native contact interactions. Overall, a small teleoperation fraction recovers performance relative to \textit{handheld} alone, but teleoperation remains the stronger data source for this task stage, highlighting the importance of capable \textit{teleoperation} interfaces.

\section{Conclusions and Limitations}
\label{sec:limitation}

In this work, we introduced \OURS, a dexterous interface for transparent teleoperation. \OURS consists of an active 7-DOF wearable thumb-to-index exoskeleton and a robot hand with identical kinematic structure. The \OURS exoskeleton can be used both in handheld data collection and leader-follower teleoperation, in which case it is capable of rendering joint-level force feedback back to the demonstrator. For both paradigms, the data collected allows training of autonomous manipulation policies executed on the follower robot. The user study, kinematic analysis, and policy learning results presented here support the following conclusions:
\begin{itemize}[leftmargin=*]
    \item \OURS matches human index-to-thumb tip-to-tip dexterity very well (96\% of poses used by seven human subjects over five tasks), and better than comparable devices. A human-like workspace is desirable both as a measure of dexterity and to mitigate constraining an operator's hand during data collection.
    \item Joint-level force feedback, enabled by the 1:1 kinematic mapping between the leader and the follower, improves both the speed and quality of teleoperated demonstrations.
    \item Data collected by \OURS allows training of effective autonomous policies that leverage intrinsic DOFs for complex tasks (flipping, inserting, manipulating an object's internal DOFs), without regrasping or large wrist motions.
    \item \OURS allows both teleoperated and handheld data collection, with force feedback in both cases, and these two data streams can be combined for efficient policy training.
\end{itemize}

The key limitations of this work span hardware and data collection. Currently, \OURS is a high-DOF two-finger device. Given the impressive capabilities of the human hand even when using only a thumb-to-index grasp, we believe that this is an avenue worthy of investigation. Nevertheless, the absence of additional fingers and a palmar surface limits the types of manipulation we can engage in. A natural extension would incorporate these components, enabling richer grasp taxonomies. Additionally, \OURS lacks tactile sensing, which is ultimately necessary for policies to make accurate contact decisions. Integrating tactile sensors remains an important direction for future work.

On the data collection side, how to best leverage the complementary nature of our two modalities remains an open question. A promising direction is to use in-the-wild collection for large-scale pretraining and teleoperation for high-quality behavior alignment. While our experiments show early evidence of the benefit of mixing these sources (Fig.~\ref{fig:itw_teleop_uncap}), understanding co-training requirements at scale, particularly for Vision-Language-Action Models, is an important avenue for future work.

\vspace{-3pt}

\bibliographystyle{IEEEtran}
\bibliography{references}

\end{document}